\documentclass[journal]{IEEEtran}

\usepackage{xcolor,soul,framed} 

\colorlet{shadecolor}{yellow}
\usepackage[pdftex]{graphicx}
\graphicspath{{../pdf/}{../eps/}}
\DeclareGraphicsExtensions{.pdf,.eps,.png}

\usepackage[cmex10]{amsmath}
\usepackage{amssymb}
\usepackage{array}
\usepackage{mdwmath}
\usepackage{mdwtab}
\usepackage{eqparbox}
\usepackage{url}
\usepackage{cite}
\usepackage{booktabs}
\usepackage{multirow}

\begin{document}
\bstctlcite{IEEEexample:BSTcontrol}
    \title{Analyzing Adversarial Robustness of Deep Neural Networks in Pixel Space: a Semantic Perspective}
  \author{Lina~Wang,~Xingshu~Chen,~Yulong~Wang,~Yawei~Yue,~Yi~Zhu,~Xuemei~Zeng,~Wei~Wang

  \thanks{L.~Wang,~X.~Chen,~Y.~Wang,~Y.~Yue~and~Y.~Zhu are with the School of Cyber Science and Engineering at Sichuan University, Chengdu 610065, China (E-mail:wlnlnw1992@163.com, chenxsh@scu.edu.cn, wangyulonga@gmail.com, yue123161@stu.scu.edu.cn) }
  \thanks{X.~Zeng~and~W.~Wang is with the Cyber Science Research Institute at Sichuan University, Chengdu 610065, China (E-mail:zengxm@scu.edu.cn, wwzqbx@hotmail.com).}}%


\maketitle

\begin{abstract}
The vulnerability of deep neural networks to adversarial examples, which are crafted maliciously by modifying the inputs with imperceptible perturbations to misled the network produce incorrect outputs, reveals the lack of robustness and poses security concerns. Previous works study the adversarial robustness of image classifiers on image level and use all the pixel information in an image indiscriminately, lacking of exploration of regions with different semantic meanings in the pixel space of an image. In this work, we fill this gap and explore the pixel space of the adversarial image by proposing an algorithm to looking for possible perturbations pixel by pixel in different regions of the segmented image. The extensive experimental results on CIFAR-10 and ImageNet verify that searching for the modified pixel in only some pixels of an image can successfully launch the one-pixel adversarial attacks without requiring all the pixels of the entire image, and there exist multiple vulnerable points scattered in different regions of an image. We also demonstrate that the adversarial robustness of different regions on the image varies with the amount of semantic information contained.
\end{abstract}


%
\IEEEpeerreviewmaketitle


\section{Introduction}

\IEEEPARstart{D}{eep} Neural Networks (DNNs) has made great advancements in recent years and has achieved impressive performance on many machine learning tasks \cite{lecun2015deep} including computer vision \cite{krizhevsky2009learning}, speech recognition \cite{vaswani2017attention}, natural language processing \cite{kumar2016ask}, to name a few. This makes DNNs a very popular technique and extensively applied in many fields. However, on the other hand, with the increasing popular of this technology, the security concerns about DNNs have become a critical topic in both academia and industry. A series of recent studies have shown that DNNs are vulnerable to a number of attacks, and the most severe and notable of which is adversarial example attack \cite{szegedy2013intriguing}. Adversarial example is generated by adding a small perturbation maliciously crafted by the attacker to the clean input and it is usually imperceptible to human. The purpose of adversarial example is to make the DNNs produce a wrong output, this threat hinders the application of deep neural networks in security-critical settings, for example, adversary may use adversarial examples to circumvent malware detection \cite{grosse2016adversarial}, fool face recognition system \cite{sharif2016accessorize} or mislead autonomous vehicle \cite{papernot2017practical}. 

Apart from the security risks, the existence of adversarial examples also presents challenges to our understanding of deep learning. It reveals the inconsistency between deep neural networks and human perception, as the adversarial examples that significantly alter the behavior of the networks are barely distinguishable from their corresponding clean inputs for human observers. Moreover, it questions the generalization ability of deep neural networks, as the capability of humans in recognizing unfamiliar objects is almost unaffected by tiny perturbations in inputs. 
Therefore, since the discovery of adversarial examples in DNNs, a body of work has been devoted to analyzing the phenomenon of adversarial examples and the adversarial robustness of deep neural networks, expecting to acquire a deeper understanding of the existing deep learning paradigm and further optimize the design of DNNs.

Previous work studies the adversarial robustness of deep neural networks at the sample level. For image classifier based on CNNs which is currently the main focus and key use case for studying the robustness of DNNs due to application popularity and security implications, the granularity of observation is an image. Some studies \cite{gu2014towards}, \cite{goodfellow2014explaining} suggest that the adversarial examples occupy large and contiguous regions in the sample space, but as far as we know, there is no research has studied the adversarial robustness in the pixel space, that is, observed the adversarial robustness at the pixel level, not at the sample level.
The advantage of humans in terms of robustness against adversarial examples compared with DNNs is believed to be due to the gap in semantic generalization ability \cite{hosseini2017limitation}, on the other hand, different parts of an image contain semantic information of varying degrees. Accordingly, it is of great significance to analyze the adversarial robustness in the pixel space of an image on the granularity of pixel. 

\begin{figure}[!htbp]
    \centering
    \includegraphics[scale=0.5]{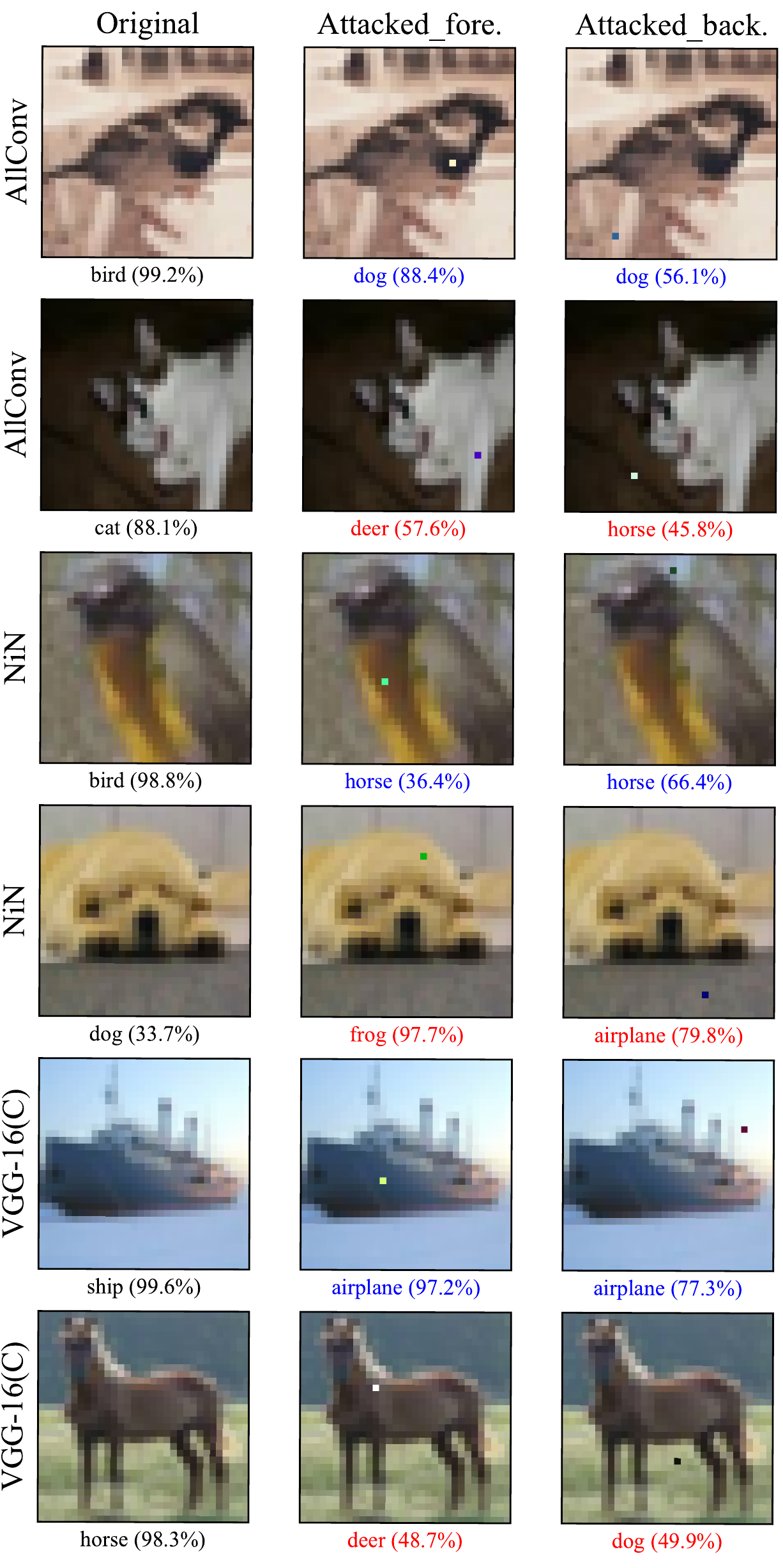}
    \caption{CIFAR-10 original images (left colomn) and the adversarial examples generated for its foreground (middle column) and background (right column) against deep neural networks of three different architectures. The original class labels are in black color while the predicted class after attacked on the foreground and the background are in blue if they are the same, red if they are different. The corresponding probability is in parentheses.}
    \label{fig:cifar_visualize}
\end{figure}

\begin{figure*}
    \centering
    \includegraphics[ width=\textwidth]{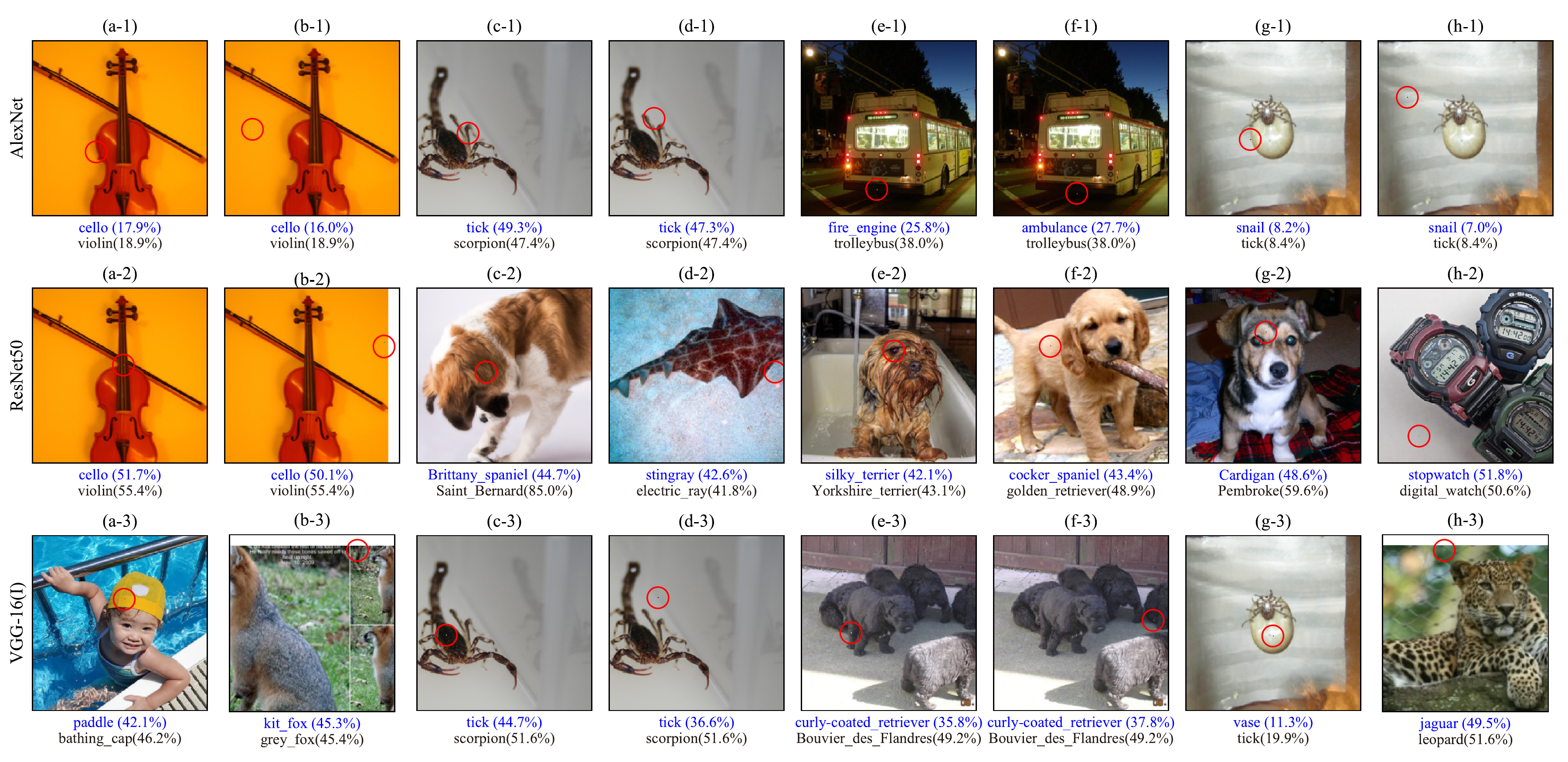}
    \caption{Illustration of the adversarial images generated by one-pixel attacks on the foreground and background of ImageNet dataset where the modified pixels are highlighted with red circles. The attacked class labels and their corresponding confidence are in blue color while the original class labels and confidence are in black below. Each row shows eight images, labeled (a$\sim$h-1/2/3), that were generated by attacks on one of the three networks: AlexNet, Resnet-50 and VGG-16(I).}
    \label{fig:imagenet_visualize}
\end{figure*}

In this paper, we fully explore the pixel space of adversarial examples by modifying only one pixel at a time on the segmented image to generate adversarial examples (see Fig.~\ref{fig:cifar_visualize} and Fig.~\ref{fig:imagenet_visualize}). In order to strictly control the adversarial perturbation occurring in a certain region of the pixel space, our research is under the framework of few-pixel attack  \cite{papernot2016limitations}, \cite{narodytska2017simple}, \cite{su2019one}, and only one pixel is modified at a time. What differs most from our work to the previous in terms of generating the few-pixel adversarial perturbation is that our algorithm only search for the perturbation in the region composed of some pixels of an image, which corresponds to an attack scenario that is harder but more realistic for the adversary since he or she is only able to have access to a part of an input image instead of the entire one. Besides, we compared the differences of adversarial robustness of deep neural networks in different regions carrying different amounts of semantic information of the image.
To the best of our knowledge, this paper is the first work to study the adversarial robustness of DNNs in pixel space with semantic information.

The contributions of this paper can be summarized as follows:
\begin{itemize}
\item We show that for successfully launching a black-box one pixel adversarial attack, it is sufficient to search for the modified pixel in some pixels of an image, rather than the pixels of the entire image. Modifying only one pixel in either the foreground region of an image or the background region can change the output label of the deep neural networks. Meanwhile, for an image, whether in its foreground or the background, the vulnerabilities are not unique, and may cause the network to misclassify the perturbed input into different target classes.  
\item We propose an algorithm to attack the deep neural networks while limiting the region where the adversarial perturbation appears on an image. The algorithm can achieve targeted or untargeted adversarial attacks in black-box scenario, it optimizes the modified individual pixel in the discontinuous intervals corresponding to the foreground or background after the image is segmented. 

\item We perform extensive experimental evaluations on a small data set CIFAR-10 \cite{krizhevsky2009learning} and a large scale data set ImageNet \cite{deng2009imagenet} to validate our algorithm and compare the differences in adversarial robustness of different regions in the pixel space. We find that the adversarial robustness of different regions on the image is related to the amount of semantic information contained in the region. The foreground region that carries important semantic information on the image is far more vulnerable to the adversarial perturbations than the background region which is almost not semantically meaningful.
\end{itemize}

The rest of this paper is organized as follows. In Section~\ref{section:2} we review the main recent works on generating adversarial examples and analyzing of adversarial examples. In Section~\ref{section:3} we present the method proposed in this paper. Experimental details, experimental results and the analysis of results are given in Section~\ref{section:4}. Section~\ref{section:5} briefly concludes the work of this paper.

\section{Related Work}
\label{section:2}
Since the work in this paper proposes a method to generate adversarial examples on the part of the image rather than the entire, and analyzes the difference of the adversarial robustness between the foreground and the background of images which contains different degree of semantic information, we summarize the work related to generating adversarial examples and analyses of adversarial vulnerability as follows.

\textbf{Generating adversarial examples.} Since Szegedy et al. \cite{szegedy2013intriguing} first revealed the vulnerability of deep neural networks, a lot of research on generating adversarial examples for the task of image classification has emerged. According to the amount of the knowledge the adversary has about the attacked model, adversarial attacks can be divided into two types: white-box attack and black-box attack.

In the white box scenario, the adversary knows some information about the model such as model type, model architecture, values of parameters, trainable weights, or has access to summary, partial, full training data. As the earliest method of generating adversarial examples, \cite{szegedy2013intriguing} used the box-constrained L-BFGS optimization method to find the smallest possible adversarial perturbation. Goodfellow et al. \cite{goodfellow2014explaining} proposed a more efficient method named FGSM that approximates the loss function of the model linearly. Kurakin et al. \cite{kurakin2016adversarial} extended the FGSM by proposing a `one-step target class' variation that uses the least likely class predicted by the network instead of the true label and a `Fast Gradient L$_\infty$' variation that uses the $L_\infty$ norm instead of the $L_2$ norm for normalization. Different from the above one-step method which only takes a single step in the direction of increasing the loss of the network to perturb images, the Basic Iterative Method (BIM) \cite{kurakin2016physical} iteratively computes the perturbation by running FGSM for several steps with a smaller step size to trade off the success rate and the computational cost. Papernot et al. \cite{papernot2016limitations} computed a saliency map to choose which pixels to modify, they used $L_0$ norm to restrict the perturbations and only altered a few pixels of the image instead of the whole image. Carlini and Wagner Attacks (C$\&$W) \cite{carlini2017towards} introduced adversarial attacks restricted on $L_0$, $L_2$ and $L_\infty$ norms, and they claimed that their attacks can make defensive distillation fail. Moosavi-Dezfooli et al. \cite{moosavi2016deepfool} proposed the DeepFool algorithm which approximates the region of the space where the network output the true label by a polyhedron and iteratively computes the perturbation vector that reaches the boundary of the polyhedron, they also provided experimental evidence to show that their method computes adversarial examples more reliably and efficiently than the existing FGSM method. Moosavi-Dezfooli et al. \cite{moosavi2017universal} also created another method to compute the image-agnostic `universal adversarial perturbations' based on DeepFool. Lina Wang et al. \cite{wang2021improving} proposed TUP method to compute targeted universal adversarial perturbations that can extract semantic information that is missing by the classifiers, and they also found that adversarial training based on TUP greatly improves robustness of deep neural networks. Baluja and Fischer \cite{baluja2018learning} used a feed-forward neural network named Adversarial Transformation Networks (ATNs) to generate adversarial examples.

In the black box scenario, the adversary is assumed to be absolutely unable to know about the model parameters, but can or cannot query the model with limited or no knowledge about the model. Many black-box attacks utilize the `transferability' property of adversarial examples, which means that adversarial examples generated for a specific model can be generalized to other models with different architectures. Papernot et al. \cite{papernot2017practical} proposed a practical black-box attack that uses a local network trained on inputs synthesized by the adversary and labels generated by the target model. Based on this practical method,  Papernot et al. \cite{papernot2016transferability} also trained the substitute model both with deep learning and techniques other than deep learning, and explore the transferability across the machine learning space. Yanpei Liu et al. \cite{DBLP:conf/iclr/LiuCLS17} studied the adversarial transferability over large models and large scale dataset, the method they proposed uses an ensemble of several models for generating adversarial examples in the complete black box scenario where the adversary cannot query the model. When the adversary can observe the outputs given by the attacked network, it becomes possible to compute adversarial perturbations using optimization algorithms that does not need to calculate the gradients. Wieland Brendel et al. \cite{brendel2018decisionbased} introduced a decision-based black-box attack named Boundary Attack that does not rely on substitute models but utilizes the final decision of the model, the Boundary Attack performs a random walk along the boundary between the adversarial and non-adversarial region from a known adversarial point to stay in the adversarial region and reduce the distance towards the target point. Chen et al. proposed Zeroth Order Optimization (ZOO) \cite{chen2017zoo} to estimate the gradients and then calculate the adversarial perturbations. Jiawei Su et al. \cite{su2019one} used differential evolution to perform one-pixel adversarial attacks. The method proposed in this paper also belongs to black-box attack, and we also launch the attacks in the extremely limited scenario where only one pixel in the image is allowed to modified. However, our approach generates adversarial examples based on image segmentation, and only searches for adversarial perturbations in a part of the image, rather than in the corresponding region of the entire image as in all the above studies (no matter in white box scenario or black box).

\textbf{Analyses of adversarial vulnerability.} Since the phenomenon of adversarial examples unveils the inherent weakness of the modern deep learning paradigm, this makes analyses of adversarial examples become a highly active research direction. However, up to now, the analyses of adversarial examples have not reached a broad consensus on many issues, especially on the existence of adversarial examples. Some viewpoints strongly deviate from each other, and some do not align with each other perfectly. Initially, the existence of adversarial examples was attributed to extreme nonlinearity of the decision boundaries induced by deep neural networks \cite{szegedy2013intriguing}. Fawzi et al. \cite{fawzi2015fundamental} introduced distinguishability measure to describe the difficulty of the classification task and they owed the phenomenon of adversarial examples to low flexibility in comparison, their results are not consistent with the initial belief about the high non-linearity of deep neural networks. Then a popular linearity hypothesis \cite{goodfellow2014explaining} that contradicts the previous idea interprets the existence of adversarial examples as a result of the deep neural models being too linear for easy optimization. Kortov and Hopfiled \cite{krotov2018dense} studied adversarial examples within the framework of Dense Associative Memory (DAM) models, and their results in line with the linearity hypothesis. Tanay and Griffin \cite{tanay2016boundary} argued that adversarial examples exists when a submanifold of sampled data  lies close to the classification boundary  under the perspective of boundary tilting, and they thought that the linear explanation of adversarial examples is not convincing. Fawzi et al. \cite{fawzi2016robustness} made a quantitative study on the curvature of the decision boundary and associated the robustness of classifiers to the curvature of decision boundaries. Rozsa et al. \cite{rozsa2018towards} presented the evolutionary stalling theory and they argued that the correctly classified samples can end up being stuck close the decision boundaries since their contributions to the loss are decreased, hence, these samples are susceptible to adversarial perturbations. While proposing a method for generating universal adversarial perturbations, Moosavi-Dezfooli et al. \cite{moosavi2017universal} also provided a conjecture about the reasons for the existence of the universal adversarial perturbations, which suggests that it is partly due to the exploitation of geometric correlations between the decision boundaries induced by the deep neural networks. They also further refined their theory in \cite{moosavi2018robustness}, and showed that deep networks are vulnerable to universal adversarial attacks when there exists a shared subspace along which the decision boundary is positively curved. Rozsa et al. \cite{rozsa2016accuracy} evaluated the robustness of different deep neural networks against various adversarial example generation approaches and found that networks with higher classification accuracies are more difficult to be attacked. Cubuk et al. \cite{cubuk2017intriguing} also argued that the adversarial accuracy is strongly correlated with clean accuracy. As far as we know, most of the research on analyzing adversarial robustness focus on the sample units except for \cite{tabacof2016exploring}, which claims that they probed the pixel space of adversarial images. However, their approach adds varying levels of noise to the image and observes the changes in the assigned labels, although it partially and roughly probes the pixel space, is still image-level analysis rather than pixel-level precise analysis. In contrast, our work really explores the pixel space by comparing the adversarial robustness against the adversarial perturbations generated at different pixels of the image.

\section{Methodology}

\label{section:3}
\subsection{Problem description}
The problem of generating adversarial examples can be formalized as an optimization problem of finding a minimum adversarial perturbation vector $\boldsymbol{r}$ with constraints. Let $\widehat{f} \colon \mathbb{R}^n \rightarrow \{1 \textellipsis K\}$ be a classifier with the $n$-dimensional input $\boldsymbol{x} = (x_1,\textellipsis,x_n) \in R^n$ and the output $\widehat{f}(\boldsymbol{x})$. Each component of $\widehat{f}(\boldsymbol{x})$ represents the probability that the input $\boldsymbol{x}$ belongs to each of the $K$ classes and therefore $\widehat{f}_{i}(\boldsymbol{x})$ is used to denote the probability of $\boldsymbol{x}$ belongs to the class $i$. The classifier assigns the label $\widehat{y}(\boldsymbol{x})= \mathop{argmax}\widehat{f}_{i}(\boldsymbol{x})$ to the input $\boldsymbol{x}$, which means that the classifier believes that $\boldsymbol{x}$ is most likely to belong to class $\widehat{y}(\boldsymbol{x})$. The adversaries are able to launch two types of attacks, the untargeted attack and targeted attack. In the case of untargeted attack, the adversarial goal is to cause the classifier to misclassify the adversarial example $\boldsymbol{x}' = \boldsymbol{x} +  \boldsymbol{r}$ into any incorrect label $\widehat{y}(\boldsymbol{x'}) \neq \widehat{y}(\boldsymbol{x})$ with a minimal perturbation $\boldsymbol{r}$. This is equivalent to finding the optimized solution for the following question:
\begin{equation}
\begin{split} 
&\min_{\boldsymbol{r}}  \quad d(\boldsymbol{x}, \boldsymbol{x+r})\\
&s.t.~\widehat{f}(\boldsymbol{x} + \boldsymbol{r}) \neq \widehat{f}(\boldsymbol{x}),
\end{split}
\label{equation:1}
\end{equation}
where $d(\cdot)$ is some distance metric to measure the similarity between $\boldsymbol{x}$ and $\boldsymbol{x'}$. For targeted attack, the purpose of which is to make the classifier classify the adversarial input $\boldsymbol{x'}$ as some specific target class $t$ incorrectly, the corresponding optimization problem can be formalized as follows:
\begin{equation}
\begin{split} 
&\min_{\boldsymbol{r}}  \quad d(\boldsymbol{x}, \boldsymbol{x+r})\\
&s.t.~\widehat{f}(\boldsymbol{x} + \boldsymbol{r}) = t.
\end{split}
\label{equation:2}
\end{equation}

The most widely-used distance metrics to quantify similarity for generating adversarial examples are $L_0$, $L_2$ and $L_\infty$ norms. In this paper, we use $L_0$ norm as all the literature in few-pixel perturbations did. To be more specific, we only modify one pixel and perform three-pixel and five-pixel attack as a comparison. The reason for doing this is because in order to more accurately  compare the attack effect of adversarial attacks in different areas of the image, we have to strictly limit the location of the perturbation. Besides, there are many possible ways to express equation \ref{equation:1} and \ref{equation:2} in a different form that is more suitable for optimization. Therefore, we apply another alternative formulation that is very commonly used and more intuitive for untargeted attack:
\begin{equation}
\begin{split} 
&\min_{\boldsymbol{r}^*}  \quad \widehat{f}_y(\boldsymbol{x} + \boldsymbol{r}) \\
&s.t.~\|\boldsymbol{r}\|_0 \leq l,
\end{split}
\label{equation:3}
\end{equation}
where $y$ represents the true class predicted by the classifier before being attacked, $l$ constrains the strength of the modification and for $L_0$ norm it denotes the number of pixels that are allowed to be modified. In the case of targeted attack, use $t$ denotes the target class that the adversary wants the classifier to misclassify as, the equation is as follows:
\begin{equation}
\begin{split} 
&\max_{\boldsymbol{r}^*}  \quad \widehat{f}_t(\boldsymbol{x} + \boldsymbol{r}) \\
&s.t.~\|\boldsymbol{r}\|_0 \leq l.
\end{split}
\label{equation:4}
\end{equation}

Previous works commonly search for the direction and the size of the modification over the entire input space while our approach partition the input space according to semantic information before generating adversarial perturbations. It means that the perturbation vector $\boldsymbol{r}$ only modify $l$ directions of $m$ ($m \textless n$) dimensions of the input $x$ with the other dimensions of $\boldsymbol{r}$ left to zeros.

\subsection{Image segmentation}
GrabCut \cite{rother2004grabcut} is a classic method that based on Graph Cut \cite{boykov2001interactive} for interactive image segmentation. GrabCut combines both texture (colour) information and edge (contrast) information, it can reduce the user interaction required to complete the segmentation while extracting foreground of good quality. The method firstly obtains a hard segmentation using iterative graph-cut optimisation and then uses border matting to deal with the problem of transparency around the object boundaries. In specific, GrabCut models the interactive segmentation task as a graph cut optimization problem just like other approached based on graphical models did. Let $\alpha = (\alpha_1, \textellipsis, \alpha_N)$ be an array of opacity values at each pixel to express the segmentation of the image ($0 \leq \alpha_n \leq 1$ ordinarily, or $\alpha_n \in \{0, 1\}$ for hard segmentation), $\theta$ be parameters that describe image foreground and background distributions. The segmentation problem can be regarded as equivalent to inferring the unknown opacity variables $\alpha$ from the given image data $\boldsymbol{z} = (z_1, \textellipsis, z_N)$ and $\theta$. The opacity value $\alpha$ can be estimated as a global minimum of the following energy function $\boldsymbol{E}$ that is defined so that its minimum corresponding to a good segmentation.
\begin{equation}
  \boldsymbol{E}(\alpha, \boldsymbol{k}, \theta,  \boldsymbol{z}) = U(\alpha, \boldsymbol{k}, \theta,  \boldsymbol{z}) + V(\alpha, \boldsymbol{z}).
\end{equation}
In order to apply to colour image, GrabCut uses color Gaussian Mixture Model (GMM) models in the data term $U$, as
\begin{equation}
    U(\alpha, \boldsymbol{k}, \theta,  \boldsymbol{z}) = \sum_{n}D(\alpha_n, k_n, \theta, z_n),
\label{equation:U}
\end{equation}
where $\boldsymbol{k}$ is a vector of the GMM component variables used to assign a unique GMM component to each pixel either from the foreground or the background model. Let $p(\cdot)$ be a Gaussian probability distribution and $\pi(\cdot)$ be the mixture weighting coefficients, so the equation~\ref{equation:U} can be rewritten as
\begin{equation}
   U(\alpha, \boldsymbol{k}, \theta,  \boldsymbol{z}) = \sum_{n}[-\log p(z_n | \alpha_n, k_n, \theta)-\log \pi(\alpha_n, k_n)]. 
\end{equation}
$V$ is the smoothness term and its contrast term is calculated based on Euclidean distance in colour space, as
\begin{equation}
    V(\alpha, \boldsymbol{z}) = \gamma \sum_{(m,n) \in \boldsymbol{C}}[\alpha_n \neq \alpha_m]exp-\beta \left \| z_{m} - z_{n} \right \|^{2}.
\end{equation}
Then the energy function $\boldsymbol{E}$ is minimized iteratively to obtain a good hard segmentation. After that, by fitting a polyline to the hard segmentation boundary, a closed contour $C$ is obtained. Now $\alpha_n$ ($n \in T_U$)is computed based on a new trimap $\{T_B, T_U, T_F\}$, where $T_U$ is the set of pixels that located in a ribbon of width $\pm w$ ($w$ is an empirically chosen parameter) pixels either side of $C$. This process is named border matting and GrabCut does this with regularisation and a dynamic programming (DP) algorithm. Foreground pixel stealing is also included to better estimate foreground pixel colours. 

U$^{2}$-Net~\cite{qin2020u2} is a deep learning based method to segment the most visually attractive objects in an image, it belongs to Salient Object Detection (SOD) which is widely used in image segmentation. U$^{2}$-Net uses a two-level nested U-structure architecture designed for SOD, which enables high resolution feature maps to be maintained at a low memory and computing cost. It nests two repeated U-Net modules instead of cascaded stacking them and comes up with a novel Residual U-block named RSU for capturing intra-stage multi-scale features. Thus the method is very flexible and easy to be adapted to different tasks with little performance loss. The three main components of U$^{2}$-Net are a six stages encoder, a five stages decoder and a saliency map fusion module which is attached with the decoder stages and the last encoder stage. Besides, U$^{2}$-Net use deep supervision in the training process and the training loss as follows:
\begin{equation}
    \mathcal{L} = \sum_{m=1}^{M}w_{side}^{m} l_{side}^{m} + w_{fuse}l_{fuse},
\end{equation}
here $l_{side}^{m}$ and $l_{fuse}$ are two loss terms, where $l_{side}^{m}$ is the loss of the side output saliency map and $l_{fuse}$ is the loss of the final fusion output saliency map. $w_{side}^{m}$ and $w_{fuse}$ are the wights of $l_{side}^{m}$ and $l_{fuse}$ respectively. Each loss term is calculated using the standard binary cross-entropy.

In this paper, GrabCut is used to segment the foreground and the background of CIFAR-10 images, and U$^{2}$-Net is used for ImageNet. U$^{2}$-Net works in a non-interactive way and it can be used to segment the ImageNet images very efficiently. On the other hand, the images of CIFAR-10 are low-resolution and the foreground and background colour distributions of which are not well separated. The distribution of these images is very different from the distribution of images used in the training of the deep learning based segmentation models, which makes the deep learning based method (such as U$^{2}$-Net) perform very poorly on CIFAR-10. GrabCut can obtain segmentations of good quality on CIFAR-10 with less user effort among the existed classical methods.  
\subsection{Optimization}
\label{section:3.3}
When we discuss the problem of generating adversarial examples from the perspective of optimization, there are some algorithms to choose from, and the solution to optimization problems is different for different methods. Since it becomes more difficult to use gradient information when we calculate the adversarial perturbation on a part of the image (foreground or background) rather than the whole, the method that does not use the gradient information is a reasonable option. Differential Evolution (DE) is one such approach, a very powerful derivative-free optimization algorithm for black-box optimization, which is about finding the minimum of a function without knowing its analytical form. DE does not require the objective function of the optimization problem to be differentiable, and its population selection mechanism preserves diversity so that it may be able to find better solutions than gradient-based methods or even other kinds of evolutionary algorithms more efficiently. These properties make DE can be used to solve complex multi-modal optimization problems, and very suitable for generating adversarial images where the networks are sometimes highly complex and non-differentiable or calculating gradient can be hardly realistic. In addition, DE was also used to generating one pixel adversarial perturbations for the whole image in \cite{su2019one}, this gives a fairer comparison of the results of our attacks that modifies one pixel in a part of the image.

When minimizing a function $f(x) \colon \mathbb{R}^n \rightarrow \mathbb{R}$, DE firstly represents the solutions as population of $P$ individuals and initializes them, where each individual is a parameter vector of $f(x)$. For each generation $G$, a population can be represented as 
\begin{equation}
    \mathit{x}_{\mathit{i}, G},\quad \mathit{i} = 1, 2, \textellipsis, P.
\end{equation}
Then an operation called mutation is used to generate new parameter vectors. A mutant vector is generated by adding the weighted difference between two different population vectors $\mathit{x}_{\mathit{r}_2, G}$ and $\mathit{x}_{\mathit{r}_3, G}$ to a third vector $\mathit{x}_{\mathit{r}_1, G}$ as
\begin{equation}
  \mathit{v}_{\mathit{i}, G+1} = \mathit{x}_{\mathit{r}_1, G} + F \cdot (\mathit{x}_{\mathit{r}_2, G} - \mathit{x}_{\mathit{r}_3, G}), 
\end{equation}
where $F \in \left [0, 2 \right ]$ is called mutation factor, $\mathit{r}_1 \neq \mathit{r}_2 \neq \mathit{r}_3\neq \mathit{i}$ and $\mathit{r}_1, \mathit{r}_2, \mathit{r}_3, \mathit{i} \in \left \{ 1, 2, \textellipsis, P \right \}$ are randomly chosen. The next step is called recombination, which aims to increase the diversity of the population. In this step, DE create a trial vector by mixing the information of the mutant with the information of the current population vector as 
\begin{equation}
 \mathit{u}_{\mathit{i}, G+1} = (\mathit{u}_{\mathit{1i}, G+1},\mathit{u}_{\mathit{2i}, G+1}, \textellipsis, \mathit{u}_{\mathit{Di}, G+1}), 
\end{equation}
where $D$ represents the dimension of the population vector and each component in $\mathit{u}_{\mathit{i}, G+1}$ can be decided by a method called crossover as follows:
\begin{equation}
\label{equation:13}
\begin{aligned}
  \mathit{u}_{\mathit{ji}, G+1} &=  \left\{\begin{array}{lr}v_{ji, G+1}&if(randb(j) \leqslant CR) \quad or \quad j = rnbr(i)
\\x_{ji, G}&if (randb(j) > CR) \quad and \quad j \neq rnbr(i),
\end{array}
\right.\\
j &= 1, 2, \textellipsis, D
\end{aligned}
\end{equation}
In (\ref{equation:13}), $CR \in \left [0, 1 \right ]$ is called crossover constant, $rnbr(i) \in 1, 2, \textellipsis, D$, and $randb(j)$ is the $j$th evaluation of a random number generator (different distribution followed corresponding to different crossover variations). Finally, DE compare the trial vector $\mathit{u}_{\mathit{i}, G+1}$ with the current population vector $\mathit{x}_{\mathit{i}, G}$. If $\mathit{u}_{\mathit{i}, G+1}$ is better measured by the fitness function $f(x)$, $\mathit{x}_{\mathit{i}, G}$ is replaced by $\mathit{u}_{\mathit{i}, G+1}$;  otherwise,  $\mathit{x}_{\mathit{i}, G}$ is preserved and $\mathit{u}_{\mathit{i}, G+1}$ discarded. The whole population will eventually converge towards the solution after many iterations.
\subsection{Method}
Let $C(\boldsymbol{x})$ be the set of pixels of an entire image $\boldsymbol{x}$, we first divide $C(\boldsymbol{x})$ into two sets $C_F(\boldsymbol{x})$ and $C_B(\boldsymbol{x})$ that satisfied
\begin{equation}
C(\boldsymbol{x}) = C_F(\boldsymbol{x}) \cup C_B(\boldsymbol{x}),\quad
C_F(\boldsymbol{x}) \cap C_B(\boldsymbol{x}) = \varnothing, 
\end{equation}
where $C_F(\boldsymbol{x})$ consists of pixels belong to the foreground of the image $\boldsymbol{x}$ and $C_B(\boldsymbol{x})$ is the set of pixels in the background. When the adversarial perturbation is optimized by DE, each candidate solution is encoded as a tuple containing five elements:
$(x, y, r, g, b)$. $x\raisebox{0mm}{-}y$ denotes the coordinates of the modified pixel and $(r, g, b)$ is the RGB value of the perturbation. In the standard DE algorithm, the individuals in the population are coded as continuous real numbers, which were originally used to solve optimization problems for continuous function. Each of these five elements can be regarded as taking values in a continuous interval when searching for the perturbation on the entire image. However, when the perturbation is optimized in the space corresponding to the the foreground or the background pixels, the interval of the coordinates $x\raisebox{0mm}{-}y$ is no longer continuous. Some methods encode individuals as discrete sequences of length $n$
\begin{equation}
    \pi = (\pi(1), \pi(2), \textellipsis, \pi(n)),
\end{equation}
and generate mutant individuals by randomly inserting and moving the sequences. The method proposed in this paper is somewhat different for that:
\begin{equation}
\begin{split}
       C_F = [a_1, a_2] \cup [a_3, a_4] \cup \textellipsis \cup [a_{n-1}, a_n],\\
    a_1 < a_2 < a_3 < a_4 < \textellipsis < a_n. 
\end{split}
\end{equation}
This means that the coordinates $x\raisebox{0mm}{-}y$ in the foreground are neither discrete nor take values on a continuous interval, but take values on multiple intervals (the same is true for the background). For simplicity, we solve this problem by making the optimization algorithm automatically constrain the position of the coordinates, instead of modifying the DE algorithm itself. Specifically, when we attack an image on its foreground, we first execute the DE algorithm for the first generation on the entire image  $\boldsymbol{x}$\ to get the trial vector $\mathit{u}_{\mathit{i}, 2}$ and the current population vector $\mathit{x}_{\mathit{i}, 1}$ in Section~\ref{section:3.3}. Then we use $\mathit{u}_{\mathit{i}, 2}$ and $\mathit{x}_{\mathit{i}, 1}$ respectively to modify the image $\boldsymbol{x}$, and get the modified images $\boldsymbol{x}{_{u}}^{'}$ and $\boldsymbol{x}{_{x}}^{'}$, just like the attack on the entire image. The difference is that when we compare the vector $\mathit{u}_{\mathit{i}, 2}$ and $\mathit{x}_{\mathit{i}, 1}$, the fitness function is computed on a new image $\bar{\boldsymbol{x}}$ and the set of pixels that make up the image as follows:
\begin{equation}
  C(\bar{\boldsymbol{x}}) = C_F(\boldsymbol{x}^{'}) \cup C_B(\boldsymbol{x}),  
\end{equation}
where $\boldsymbol{x}^{'}$ represents the modified images $\boldsymbol{x}{_{u}}^{'}$ or $\boldsymbol{x}{_{x}}^{'}$. Similarly, when attacking on the background, 
\begin{equation}
  C(\bar{\boldsymbol{x}}) = C_F(\boldsymbol{x}) \cup C_B(\boldsymbol{x}^{'}). 
\end{equation}
Thus, at the replacement stage of each generation, we evaluate the vector $\bar{\boldsymbol{x}}$ with the fitness function. This can ensure that the adversarial perturbation is in the segmented foreground or background, and force the DE algorithm to find the optimal solution in the corresponding area. 

It is worth noting that when we looking for perturbations in the foreground (or background) we also used pixels of the background (or foreground) at the beginning, but from the second generation the DE algorithm actually only looks for the solution in the foreground (or background) instead of the entire image. This can be confirmed by observing the change of the value of the fitness function over the generations during evolution, because when attacking on the foreground (or background), if the perturbed pixel is located in the background (or foreground), the value of the fitness function will not change at all in our method, but the actual experimental results show that the value of the fitness function drops rapidly from the beginning. 
Here we only show the feasibility of using only part of the pixels (foreground or background pixels) of an image for a successful one-pixel attack and leave the specific implementation for future work.

\section{Evaluation}
\label{section:4}

\subsection{Experimental setup}
\label{section:4.1}
In this section, we first introduce the datasets and the deep learning models used for the experimental investigation. The details of the experiment are described afterwards, and the experimental results are analyzed in depth at the end of this section. 

\textbf{Dataset description.} The experiments were performed on two datasets of different scale, CIFAR-10 and ImageNet, both of which are widely used. The two datasets contain a different number of color images and each pixel of these images takes the value of a real number between $0$ and $255$ for three color channels. Specifically, the CIFAR-10 dataset is a collection of tiny images in $10$ classes which contains $50,000$  training images and $10,000$ test images, each with a resolution of $32 \times 32$. The ImageNet project is a large visual database and ILSVRC 2012 \cite{ILSVRC15}, which uses a subset of ImageNet with $1.2$ million training images, $50,000$ validation images and $150,000$ test images in $1000$ categories, is one of the most commonly used version of ImageNet. For each of the attacks on the three types of deep neural networks trained on CIFAR-10, we randomly selected $500$ images from CIFAR-10 test dataset. Analogously, for ImageNet, $420$ images from ILSVRC 2012 test set were randomly selected for each of the attacks. Besides, we also resized the ImageNet images to $224 \times 224$ resolution since the original ILSVRC 2012 images are not uniform in size.

\textbf{Architecture characteristics.} In order to conduct the adversarial attacks, we trained three of common networks on CIFAR-10: all convolutional network (AllConv) \cite{DB15a}; network in network (NiN) \cite{lin2014network} and VGG-16 network \cite{DBLP:journals/corr/SimonyanZ14a}. And we also trained three networks on ILSVRC 2012: BVLC AlexNet \cite{krizhevsky2012imagenet}; Residual Network with 50 layers (ResNet-50) \cite{he2016deep} and VGG-16. We trained the network VGG-16 on both the two datasets for comparing the performance of the same architecture on different data sets. To distinguish between the two, we use VGG-16 (C) represent the VGG-16 network trained on CIFAR-10 and use VGG-16 (I) represent the network trained on ImageNet in this paper. We will also omit the letters in parentheses when there is no ambiguity. The architectures of the networks follow those described in the original works. The baseline accuracy of all networks on clean test sets is shown by Table~\ref{tabel:1} which is comparable to state-of-the-art performance.
\begin{table}[h]
\caption{Baseline accuracy (acc.) of three CIFAR-10 networks and three ImageNet networks.}
\label{tabel:1}
\begin{tabular}{m{0.25\textwidth} m{0.08\textwidth} m{0.08\textwidth}}
\hline
\textbf{Architecture (CIFAR-10)}        & \multicolumn{2}{c}{\textbf{Acc. (\%) }}  \\ \hline
AllConv     & \multicolumn{2}{c}{85.71} \\
NiN & \multicolumn{2}{c}{86.55} \\
VGG-16 (C) & \multicolumn{2}{c}{85.65} \\\hline
\textbf{Architecture (ImageNet)}     & \textbf{Top1 (\%) }  & \textbf{Top5 (\%) }  \\ \hline
AlexNet    & 56.52  & 79.07\\
ResNet-50 & 75.90  & 92.90\\
VGG-16 (I) & 71.50  & 90.10\\\hline
\end{tabular}
\end{table}

\subsection{Implementation details}
\label{section:4.2}
\textbf{CIFAR-10.} We launched untargeted adversarial attacks and targeted attacks on $500$ randomly selected CIFAR-10 test images for the three CIFAR-10 neural networks. For untargeted attacks, if an image can be perturbed to be misclassified as any other class, the untargeted attack on this image succeeds. For targeted attacks, each image is perturbed to all other nine CIFAR-10 classes separately, and we count it as a successful targeted attack only if the image is perturbed to the specified target class. That is, when an image classified as class $A$ without being attacked is to be perturbed to a target class $B$, but it is finally misclassified as another target class $C \neq B \neq A$, this cannot be counted as a successful targeted attack. 

\begin{table*}[htp]
\centering
\caption{Results of conducting one-, three-, five-pixel attacks in CIFAR-10 original test images, the foreground and background of the images on three types of networks: AllConv, NiN, VGG-16. The Untargeted1/Targeted indicate the success rate of conducting untargeted/targeted attacks. Untargeted2 is the success rate calculated from targeted attack results. Confidence$_{u}$/ Confidence$_{t}$ is the average probability of successful untargeted/targeted attacks.} 
\label{tabel:2}
\begin{tabular}{lllllllllll}
\toprule
\multirow{2}{*}{\textbf{Networks}}          &       \multirow{2}{*}{\textbf{Metric}}        & \multicolumn{3}{c}{\textbf{Original}} & \multicolumn{3}{c}{\textbf{Foreground}} & \multicolumn{3}{c}{\textbf{Background}} \\ \cline{3-11} 
                           &      & \textbf{1-pixel}        & \textbf{3-pixels}       & \textbf{5-pixels}       & \textbf{1-pixel}         & \textbf{3-pixels}       & \textbf{5-pixels}       & \textbf{1-pixel}         & \textbf{3-pixels}        & \textbf{5-pixels}        \\ \hline
\multirow{5}{*}{\textbf{AllConv}}   & Untargeted1 & 12.80\%  & 47.00\% & 51.20\% & 11.80\%  & 40.20\%  & 41.20\%  & 6.20\%   & 19.40\%  & 20.60\%  \\ \cline{2-11} 
                           & Confidence$_u$  & 74.01\%  & 70.12\% & 75.35\% & 71.38\%  & 72.75\%  & 77.13\%  & 74.71\%  & 72.32\%  & 80.09\%  \\ \cline{2-11} 
                           & Untargeted2 & 12.60\%  & 48.20\% & 55.00\% & 11.80\%  & 43.60\%  & 45.80\%  & 6.60\%   & 22.20\%  & 24.00\%  \\ \cline{2-11} 
                           & Targeted   & 2.93\%   & 13.11\% & 16.84\% & 2.91\%   & 11.42\%  & 13.27\%  & 1.78\%   & 5.51\%   & 6.84\%   \\ \cline{2-11} 
                           & Confidence$_t$  & 67.51\%  & 61.75\% & 65.99\% & 66.72\%  & 64.05\%  & 68.58\%  & 68.09\%  & 66.51\%  & 72.07\%  \\ \midrule[1pt]
\multirow{5}{*}{\textbf{NiN}}       & Untargeted1 & 21.00\%  & 67.80\% & 75.40\% & 19.20\%  & 59.80\%  & 60.00\%  & 9.80\%   & 35.60\%  & 35.20\%  \\ \cline{2-11} 
                           & Confidence$_u$  & 66.40\%  & 63.77\% & 66.23\% & 63.67\%  & 64.25\%  & 69.29\%  & 64.14\%  & 64.65\%  & 70.14\%  \\ \cline{2-11} 
                           & Untargeted2 & 21.00\%  & 69.60\% & 76.80\% & 19.40\%  & 64.00\%  & 69.20\%  & 11.40\%  & 40.80\%  & 45.00\%  \\ \cline{2-11} 
                           & Targeted    & 5.91\%   & 30.62\% & 38.56\% & 5.56\%   & 25.07\%  & 28.49\%  & 2.71\%   & 13.84\%  & 16.07\%  \\ \cline{2-11} 
                           & Confidence$_t$  & 59.50\%  & 53.41\% & 54.40\% & 59.57\%  & 54.71\%  & 57.72\%  & 59.79\%  & 56.41\%  & 60.70\%  \\ \midrule[1pt]
\multirow{5}{*}{\textbf{VGG-16 (C)}} & Untargeted1 & 43.40\%  & 88.40\% & 92.00\% & 41.60\%  & 79.60\%  & 79.00\%  & 18.20\%  & 51.40\%  & 48.20\%  \\ \cline{2-11} 
                           & Confidence$_u$  & 70.06\%  & 69.98\% & 73.80\% & 70.00\%  & 71.62\%  & 74.82\%  & 67.52\%  & 68.36\%  & 72.40\%  \\ \cline{2-11} 
                           & Untargeted2 & 44.60\%  & 90.80\% & 93.20\% & 42.80\%  & 87.00\%  & 90.60\%  & 20.20\%  & 61.60\%  & 66.80\%  \\ \cline{2-11} 
                           & Targeted    & 22.78\%  & 68.13\% & 70.80\% & 22.33\%  & 61.24\%  & 60.64\%  & 8.82\%   & 35.04\%  & 36.80\%  \\ \cline{2-11} 
                           & Confidence$_t$  & 59.38\%  & 51.55\% & 53.66\% & 59.94\%  & 53.39\%  & 57.85\%  & 58.40\%  & 54.54\%  & 59.26\%  \\ \bottomrule
\end{tabular}
\end{table*}

Then we segmented the foreground and the background of these test images use GrabCut. The interaction with GrabCut (the amount of which is usually small) stops until a satisfactory result for human eyes is obtained. The untargeted and targeted adversarial attacks using the same technique were conducted in the foreground and in the background. In order to compare the effects of the adversarial attack in the foreground and the background, it is necessary to strictly control the position where the adversarial perturbation is added. We thus performed the attacks in an extremely limited scenario where the perturbation only can be added to one pixel. In addition, we also conducted the experiments by perturbing the images with three and five pixel-modification in the original natural images as well as their foreground and background on all the three CIFAR-10 networks. The purpose of this is to explore the situation of different perturbation scale. 

The perturbation used to modify the pixel (or pixels) is computed by the population-based optimization algorithm DE. The initial number of population is set to $400$ and a stratified sampling method called \emph{latin hypercube sampling} (LHS) is used to initialize the initial population to maximize coverage of the available parameter space. The maximum number of iteration is set to $100$ and early-stop criterion will be triggered when the predicted label is no longer the true class in the case of untargeted attacks, and when the predicted label is the target class in the case of targeted attacks. We also employ dithering when set the value of mutation factor $F$ to help speed convergence significantly, it means $F$ takes a uniform random number between $[0.5, 1)$ in each iteration. The binomial crossover is included. The fitness function is the confidence value of true class for untargeted attack and the confidence value of target class for targeted attack.

\textbf{ImageNet.} We only performed one-pixel untargeted attacks on three ImageNet neural networks with the same DE parameter settings considering the time constraints. Although the search space of ImageNet is much larger than that of CIFAR-10 (approximately $50$ times larger), it is enough to see a clear trend from the experimental results even without proportionally increasing the number of evaluations. This low computational cost setting corresponds to a computationally tractable attack method on ImageNet, which is more realistic. Of cause, we also believe that more evaluations may result in a higher attack success rate.

In addition, the most different between the experiment on ImageNet and CIGAR-10 is that we used a different method called U$^{2}$-Net to segment the foreground and background from ImageNet natural images. The setting of U$^{2}$-Net was kept as similar as possible to the original with a few modifications to adapt to changes in input size. To ensure that the foreground and background of each image used for attack are properly segmented, we finally performed a manual inspection as an additional guarantee.

\subsection{Experimental results}
\label{section:4.3}
Some examples of perturbed images are visualized in Fig.~\ref{fig:cifar_visualize} and Fig.~\ref{fig:imagenet_visualize}. It can be clearly seen that vulnerable points in the pixel space corresponding to an image are diverse and not unique. For the same architecture trained on the same data set, the perturbed images obtained by modifying one pixel in the foreground and modifying one pixel in the background may be misclassified by the model into the same class (see the images with labels in blue in Fig.~\ref{fig:cifar_visualize}, or (a-1) and (b-1), (c-1) and (d-1), (a-2) and (b-2), (c-3) and (d-3), etc in Fig.~\ref{fig:imagenet_visualize}) or different classes (images with labels in red in Fig.~\ref{fig:cifar_visualize}, (e-1) and (f-1) in Fig.~\ref{fig:imagenet_visualize}). For models with different architectures trained on the same data set, there is more than one candidate pixel both in the foreground and background when performing the one-pixel attack, which makes different models incorrectly classify the perturbed images into the same class (see (a-1) and (a-2), (b-1) and (b-2), (c-1) and (c-3), (d-1) and (d-3) in  Fig.~\ref{fig:imagenet_visualize}) or different classes ((g-1) and (g-3) in Fig.~\ref{fig:imagenet_visualize})). 

\textbf{Success rate.} In order to evaluate the attack performance on the foreground and background of the image, we introduce several metrics to measure the effectiveness of the attacks, with only a few differences for CIFAR-10 and ImageNet datasets. 
Specifically, we report the success rate of untargeted attacks and targeted attacks as well as their corresponding confidence for CIFAR-10 in Table~\ref{tabel:2}. The success rate is defined as the percentage of adversarial examples that are successfully misclassified as any incorrect class label in the case of untargeted attacks. And in the case of targeted attacks, it is defined as the probability  of adversarial examples being misclassified into a specific target class. In addition to the success rate calculated based on the actual untargeted and targeted attack results, we also report the untargeted success rate evaluated based on targeted attack results. That is to say, if the targeted attack can be launched successfully on an image at least for one target class, the untargeted attack on this image is considered successful. When calculating the confidence value, we accumulate the values of probability of the predicted class after each successful attack, then divided by the number of successful attacks. For ImageNet, we report the success rate and the corresponding confidence of top1 predicted class after being attacked in the case of untargeted attack. Besides, since we focus on decreasing the probability of the original true class, in order to further investigate the effect of the attacks, the average decrease in confidence of the original top1, top3, top5 classes before and after being attacked as well as the probability that the original top1, top3, top5 classes is still in top3, top5 prediction after being attacked are also shown in Table~\ref{tabel:3}.
\begin{table*}[htp]
\centering
\caption{Results of conducting one-pixel untargeted attack in Imagenet original test images on three types of networks: Alexnet, VGG-16, ResNet-50. The Untargeted and Confidence$_u$ represent the top-1 attack success rate and the average confidence of the images successfully attacked respectively. The Conf.-Decrease-top1/3/5(succ/unsucc) indicate the average decrease in confidence of the original top1/3/5 classes predicted by the unattacked classifiers before and after successful/unsuccessful attacks. The Ori-top1/3/5-in-top3/5(succ/unsucc) indicate the probability that the original top1/3/5 predicted class is still in the top3/5 after being attacked successful/unsuccessfully.} 
\label{tabel:3}
\setlength{\tabcolsep}{7mm}{
\begin{tabular}{lllll}
\toprule
\textbf{Networks}                    & \textbf{Metric} & \textbf{Original} & \textbf{Foreground} & \textbf{Background} \\ \hline
\multirow{9}{*}{\textbf{Alexnet}}    & Untargeted                          & 11.19\%                               & 10.71\%                                 & 6.90\%                                  \\ \cline{2-5} 
                                     & Confidence$_u$                       & 30.68\%                               & 31.14\%                                 & 29.06\%                                 \\ \cline{2-5} 
                                     & Conf.-Decrease-top1(succ)         & 11.41\%                               & 11.94\%                                 & 7.09\%                                  \\ \cline{2-5} 
                                     & Conf.-Decrease-top3(unsucc)       & 0.76\%                                & 0.71\%                                  & 0.52\%                                  \\ \cline{2-5} 
                                     & Conf.-Decrease-top5(unsucc)       & 0.32\%                                & 0.29\%                                  & 0.21\%                                  \\ \cline{2-5} 
                                     & Ori-top1-in-top3(succ)           & 100.00\%                              & 97.78\%                                 & 100.00\%                                \\ \cline{2-5} 
                                     & Ori-top1-in-top5(succ)           & 100.00\%                              & 100.00\%                                & 100.00\%                                \\ \cline{2-5} 
                                     & Ori-top3-in-top3(unsucc)         & 92.58\%                               & 92.53\%                                 & 95.74\%                                 \\ \cline{2-5} 
                                     & Ori-top5-in-top5(unsucc)         & 92.60\%                               & 92.85\%                                 & 95.19\%                                 \\ \midrule[1pt]
\multirow{9}{*}{\textbf{VGG-16 (I)}} & Untargeted                          & 3.81\%                                & 3.81\%                                  & 1.43\%                                  \\ \cline{2-5} 
                                     & Confidence$_u$                       & 38.12\%                               & 38.40\%                                 & 42.32\%                                 \\ \cline{2-5} 
                                     & Conf.-Decrease-top1(succ)         & 13.83\%                               & 14.45\%                                 & 8.65\%                                  \\ \cline{2-5} 
                                     & Conf.-Decrease-top3(unsucc)       & 0.39\%                                & 0.37\%                                  & 0.23\%                                  \\ \cline{2-5} 
                                     & Conf.-Decrease-top5(unsucc)       & 0.16\%                                & 0.15\%                                  & 0.10\%                                  \\ \cline{2-5} 
                                     & Ori-top1-in-top3(succ)           & 100.00\%                              & 100.00\%                                & 100.00\%                                \\ \cline{2-5} 
                                     & Ori-top1-in-top5(succ)           & 100.00\%                              & 100.00\%                                & 100.00\%                                \\ \cline{2-5} 
                                     & Ori-top3-in-top3(unsucc)         & 95.06\%                               & 95.09\%                                 & 96.96\%                                 \\ \cline{2-5} 
                                     & Ori-top5-in-top5(unsucc)         & 95.51\%                               & 95.48\%                                 & 96.92\%                                 \\ \midrule[1pt]
\multirow{9}{*}{\textbf{ResNet-50}}  & Untargeted                          & 3.57\%                                & 3.10\%                                  & 0.95\%                                  \\ \cline{2-5} 
                                     & Confidence$_u$                       & 44.92\%                               & 43.78\%                                 & 45.44\%                                 \\ \cline{2-5} 
                                     & Conf.-Decrease-top1(succ)         & 21.29\%                               & 19.23\%                                 & 11.02\%                                 \\ \cline{2-5} 
                                     & Conf.-Decrease-top3(unsucc)       & 0.38\%                                & 0.37\%                                  & 0.21\%                                  \\ \cline{2-5} 
                                     & Conf.-Decrease-top5(unsucc)       & 0.13\%                                & 0.13\%                                  & 0.08\%                                  \\ \cline{2-5} 
                                     & Ori-top1-in-top3(succ)           & 100.00\%                              & 100.00\%                                & 100.00\%                                \\ \cline{2-5} 
                                     & Ori-top1-in-top5(succ)           & 100.00\%                              & 100.00\%                                & 100.00\%                                \\ \cline{2-5} 
                                     & Ori-top3-in-top3(unsucc)         & 93.40\%                               & 93.32\%                                 & 95.17\%                                 \\ \cline{2-5} 
                                     & Ori-top5-in-top5(unsucc)         & 93.81\%                               & 94.01\%                                 & 96.71\%                                 \\ \bottomrule
\end{tabular}}
\end{table*}

On CIFAR-10, the results of one-pixel untargeted attacks and targeted attacks both show that the attack performance on the foreground of the image is comparable to the attack effect on the original entire image, while the attack success rate on the background is significantly lower. For example, there is $43.40\%$ chance that the one-pixel untargeted attack on an arbitrary CIFAR-10 original test image can cause the image to be misclassified for model VGG-16 (C), $41.60\%$ chance that on the foreground of the image the attack can success, while the success rate drops to $18.20\%$ for the attack on the background of the image. For all the three CIFAR-10 models, the difference between the success rate on the original image and the success rate on its background is much larger than the difference between the success rate on the original image and the success rate on its foreground. The success rate of the three models on the foreground of the image reduces by an average of $6.84\%$ in the case of untargeted attacks and by an average of $2.86\%$ in the case of targeted attacks compared with the success rate on the original image. In contrast, the success rate on the background of the image decreases by an average of $54.32\%$ for untargeted attacks and $51.56\%$ for targeted attacks respectively compared with the success rate on the original image. Moreover, for stronger attacks obtained by increasing the number of pixels that can be modified to three and five, the value of success rate increases significantly, while the observation about the success rate of the attacks on the original image compared with its foreground and background has not changed. Besides, for a successful untargeted attack or targeted attack, attacking on the original image or its foreground or background does not significantly change the confidence value corresponding to the predicted class after being attacked, nor does the confidence value change with the number of pixels that can be modified.

On ImageNet, we observed that the success rate of the three models on the foreground of the image decreases by an average of only $5.86\%$ compared with the success rate on the original image, while the success rate on the background reduces by an average of $58.04\%$ compared with on the original image, which is almost 10-fold greater. Regardless of whether the modified pixels are located in the entire image or its foreground or the background, the confidence value of the predicted class after being attacked successfully does not change much. These results are in line with our previous finding of CIFAR-10 and show that the observation generalizes well to large size images. For successful attacks, the average of the reduction in confidence corresponding to the original true top1 class of the three ImageNet models are, $15.51\%$ for attacks on the original image, $15.21\%$ for attacks on the foreground of the image and $8.92\%$ for attacks on the background respectively. Even for the unsuccessful attacks, the drop in confidence values corresponding to the true top3 and top5 classes after being attacked on the background of the image is smaller than that after being attacked on the entire image and its foreground for all the three models. All these results tell us that attacking on the background of the image is much less effective than on the entire image or its foreground. But it should also be noted that, although the average probability that the original top3 or top5 classes are still in the top3 or top5 predictions is highest after being unsuccessfully attacked on the background than on the foreground and the entire image, the original top3 or top5 classes are still in the top3 or top5 predictions in more than $90\%$ of the attacks for all the three models. If the attacks are launched successfully, the probability that the original top1 class is still in top3 or top5 predictions is almost $100\%$. This small change in the rank order of the predicted classes is caused by the fact that we only utilized a simple implementation to modify the pixel of the images for such a large scale dataset. More calculations, other setting of parameters and fitness functions should give different results.
\begin{figure*}
    \centering
    \includegraphics[scale=0.2, width=7.0in]{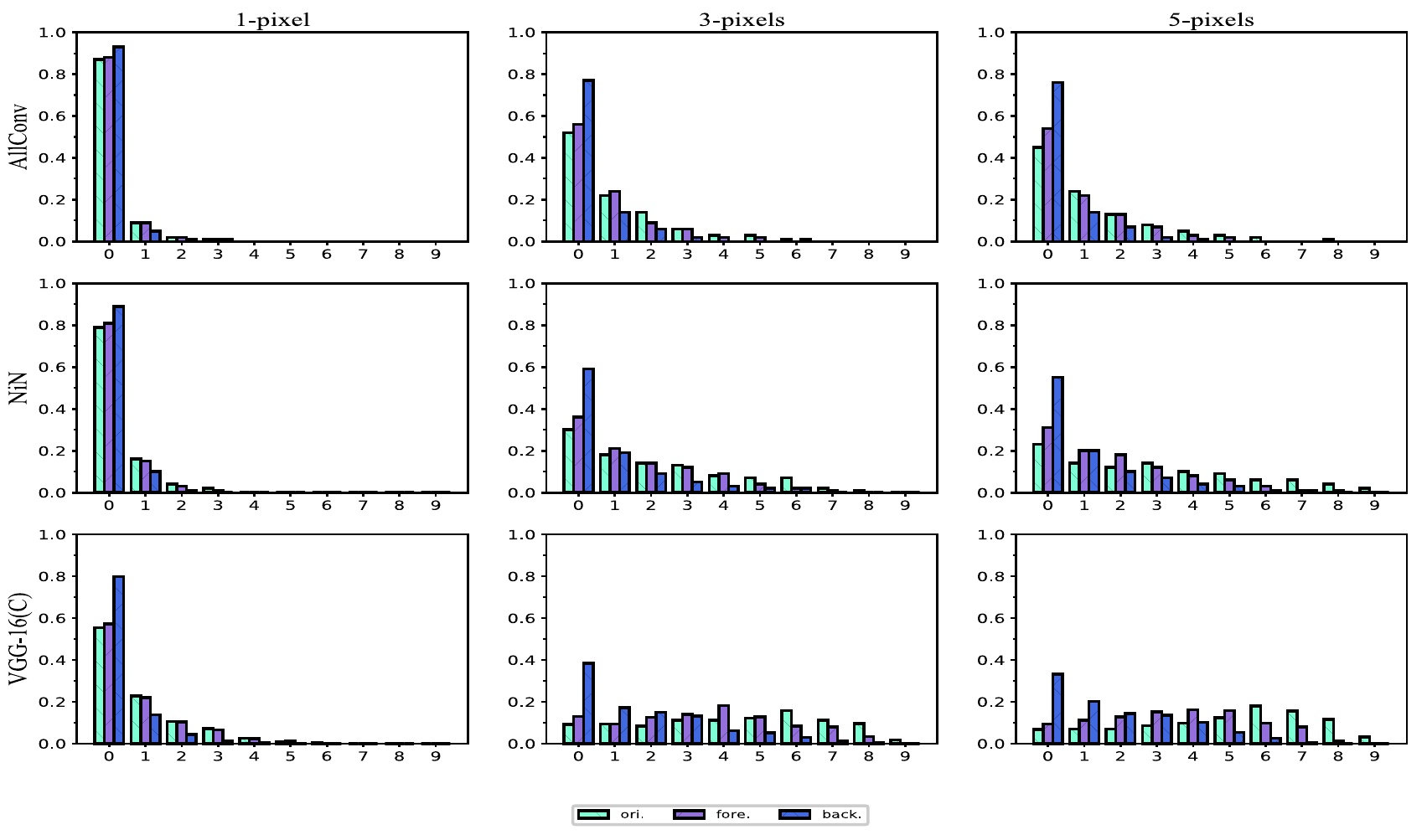}
    \caption{Percentage of CIFAR-10 test images that were successfully perturbed to a certain number (from $0$ to $9$) of target classes by one-, three- five-pixel attacks on the  original entire images, its foreground and background}.
    \label{fig:num_success}
\end{figure*}

\textbf{Number of target classes (successful targeted attack results).} The number of target classes that the percentage of the images can be perturbed to are shown in Fig.~\ref{fig:num_success} where one-, three-, five-pixels are modified on the entire image, its foreground and background respectively. This metric is only included in the CIFAR-10 results since there are too many classes of ImageNet classification up to $1000$ classes. It can be seen that with only one-pixel modification on the entire image, a certain amount of images can be misclassified up to three target classes by the AllConv network and NiN network, and a small number of images can even be perturbed to six target classes by VGG-16 (C) network. The results of the number of target classes when attacking on the entire image are almost as good as the results when attacking on the foreground of the image, whereas the results of the attacks that modify the pixel on the background of the image are much worse.

Perturbing the images to more target classes can be achieved when increasing the number of pixels that can be modified. At the same time, there is still a large gap between the attack effect on the background of the image and that in the entire image or its foreground. These results suggest that the parts of an image that contain different amounts of semantic information are vulnerable to adversarial attacks to different degrees.

\begin{figure}[ht]
    \centering
    \includegraphics[width=0.5\textwidth]{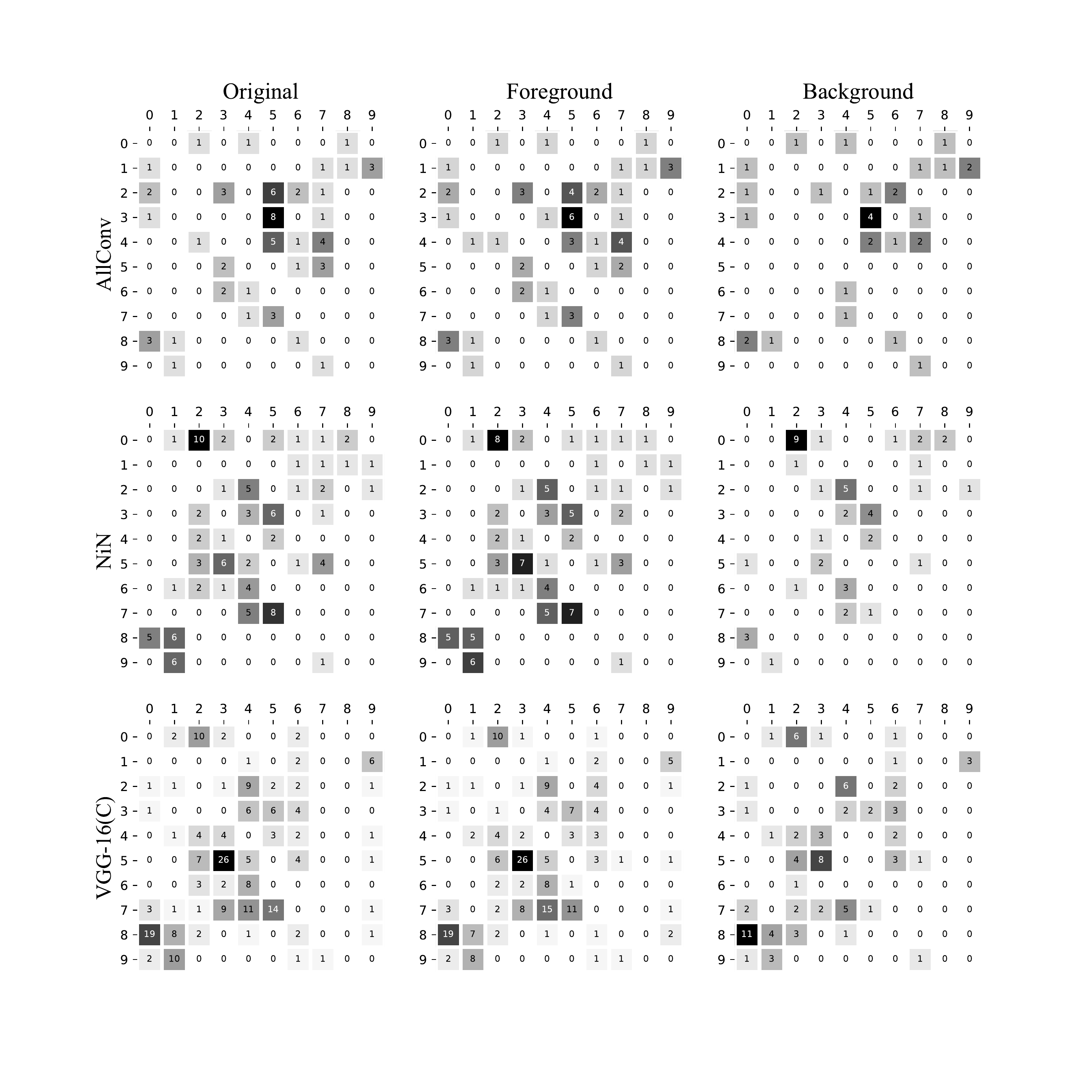}
    \caption{Heat maps of the number of times the untargeted one-pixel attack was successful with the corresponding source–target class pair, for the original, foreground and background images of CIFAR-10 on the three type of networks: AllConv, NiN and VGG-16(C). The vertical indices indicate the original classes and the horizontal indices indicate the target classes. The number from $0$ to $9$ in the vertical and horizontal indices represent the ten classes of CIFAR-10: airplane, automobile, bird, cat, deer, dog, frog, horse, ship, and truck.}.
    \label{fig:heatmap_un}
\end{figure}

\begin{figure}
    \centering
    \includegraphics[width=0.5\textwidth]{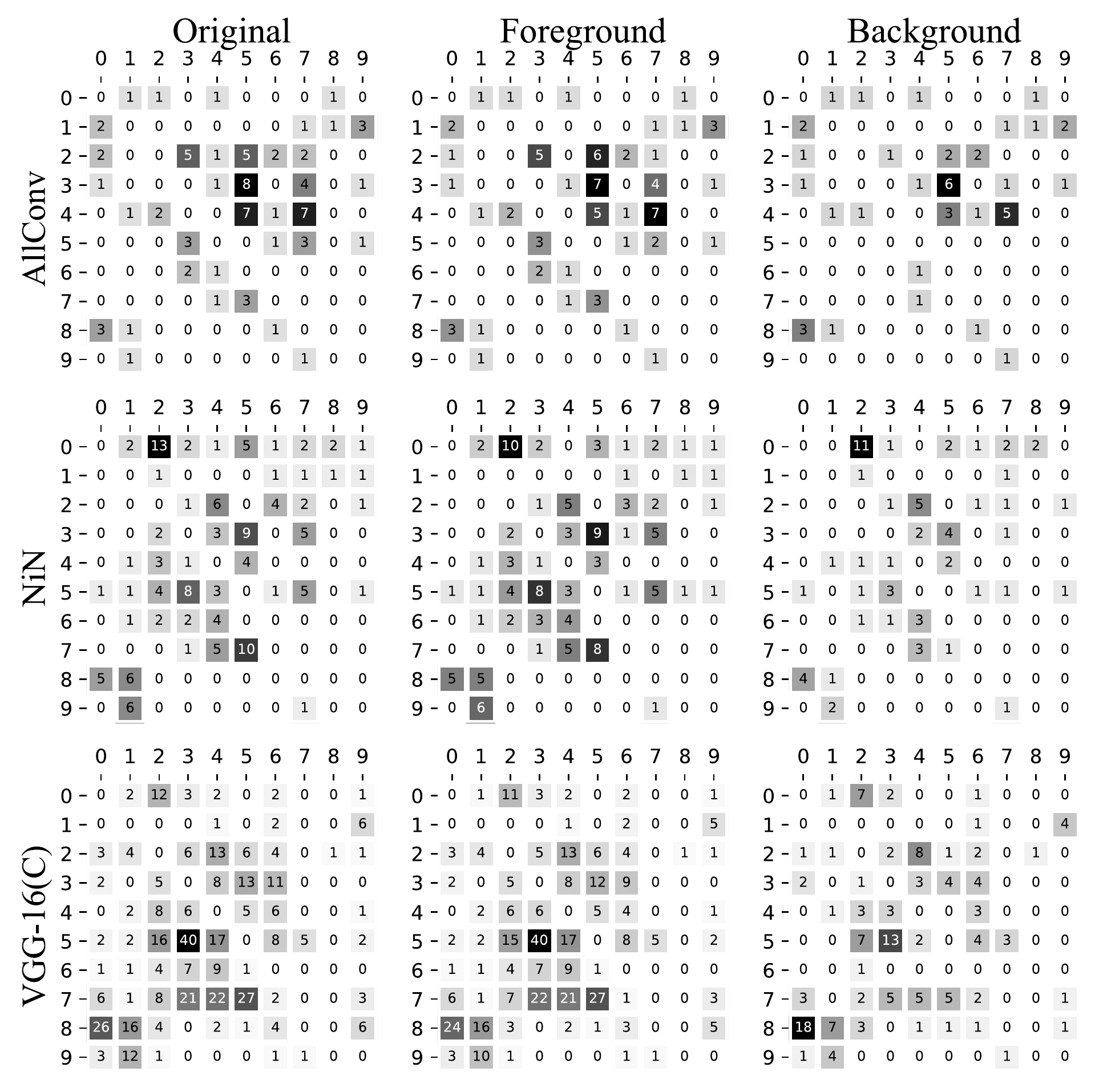}
    \caption{Heat maps of the number of times the targeted one-pixel attack was successful with the corresponding source–target class pair, for the original, foreground and background images of CIFAR-10 on the three type of networks: AllConv, NiN and VGG-16(C). The vertical indices indicate the original classes and the horizontal indices indicate the target classes. The number from $0$ to $9$ in the vertical and horizontal indices represent the ten classes of CIFAR-10: airplane, automobile, bird, cat, deer, dog, frog, horse, ship, and truck.}.
    \label{fig:heatmap_tar}
\end{figure}

\textbf{Original-target class pairs.} We display the number of the successful untargeted and targeted one-pixel attacks corresponding to different original-target class pairs in Fig.~\ref{fig:heatmap_un} and Fig.~\ref{fig:heatmap_tar}. The heat-maps in Fig.~\ref{fig:heatmap_un} and Fig.~\ref{fig:heatmap_tar} are based on the results of the three CIFAR-10 networks when attacking on the original entire image as well as on its foreground and background. It can be observed that the results of untargeted and targeted attacks show no significant difference in patterns. When attacking on the entire image, the degree of vulnerability varies for different original-target class pairs and there are some specific original-target class pairs which are much more vulnerable than others. For example, images of class $5$ can be much more easily misclassified to class $3$ but can hardly be perturbed to class $8$ for the targeted attacks on the VGG-16 (C) network. Most of these specific class pairs are still more vulnerable than others to attacks on the foreground of the image, but are no longer significantly different from other class pairs when attacking on the background. It suggests that for a single image, the vulnerable directions are more likely found in the space corresponding to the part of the image that contains more semantic information, such as the foreground. 

In addition, it is more difficult to be attacked successfully for the images of some classes than of others (such as class $9$ for AllConv and NiN, class $1$ for VGG-16 (C)), and some classes are harder to reach (such as class $2$ for AllConv, class $9$ for NiN, class $8$ for VGG-16 (C)). For the classes that are hard to be attacked and to be perturbed to, the difficulty is further increased when attacking on the background of the image. 
This indicates that the vulnerable directions shared by different data points that belong to the same class are also closely related to the amount of semantic information. It is more difficult to search for the vulnerable directions on the background, which contains less semantic information than the foreground of the image. Therefore, when the data points going across the input space through the direction parallel to any of the directions corresponding to the pixels in the background, the original classes are more likely to keep robust along these directions. Meanwhile, for the classes that are already hard to reach, finding a reachable direction in the space corresponding to the background will be even more difficult. This phenomenon might be due to the shape of the decision boundary and the relative angle between each direction of the data point and the boundary. It means that, for a wide boundary that the data points may be far away from the boundary, the distance from the boundary may be further along the directions in the space corresponding to the background. On the other hand, if the boundary shape is thin, it is more difficult to reach along the direction corresponding to the pixel with less semantic information.

\begin{figure*}
    \centering
    \includegraphics[scale=0.6, width=7.0in]{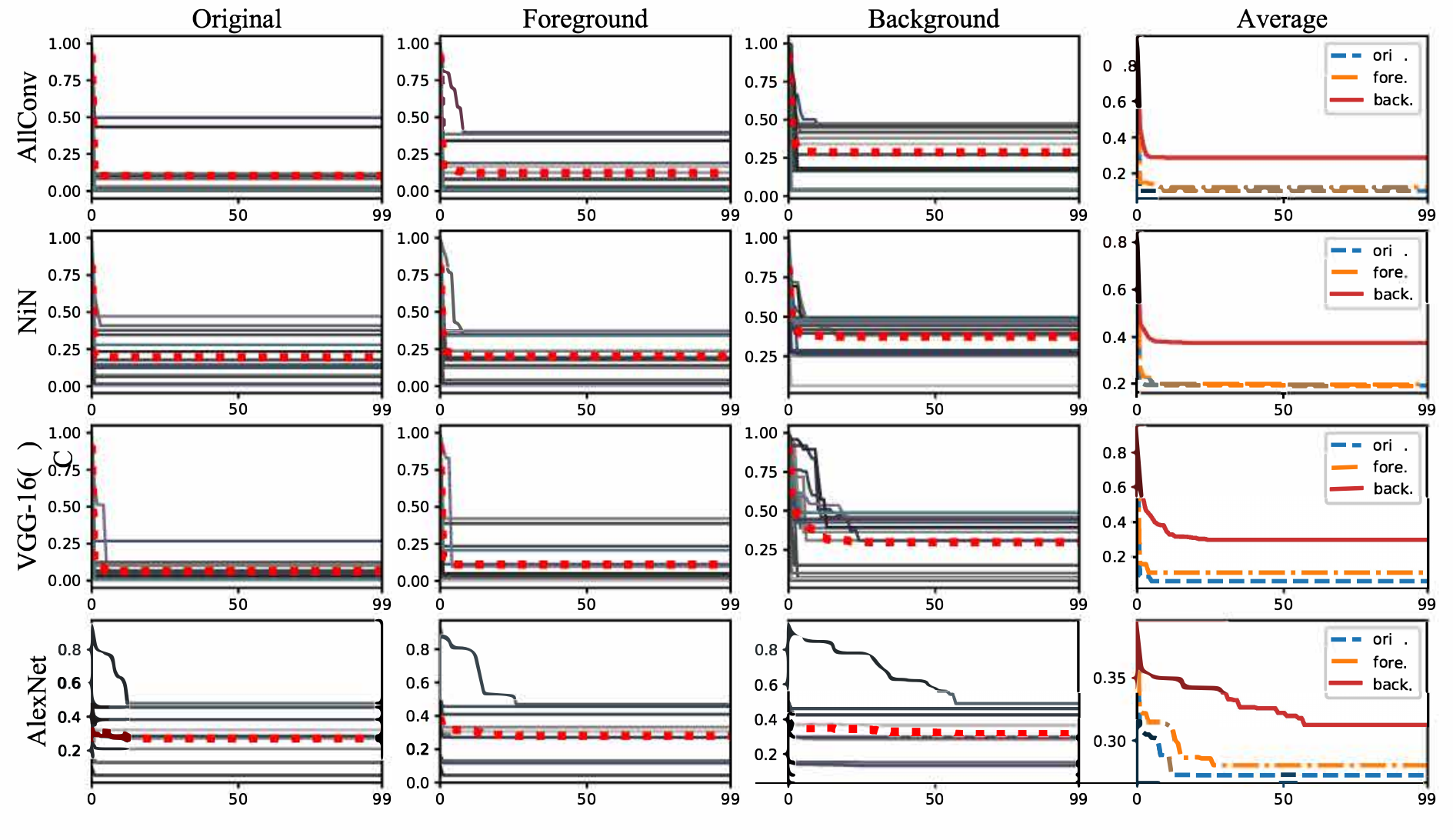}
    \caption{Changes of fitness values during $100$ generations of evolution of untargeted attacks by modifying one pixel on the entire image and its foreground and background among the three CIFAR-10 networks: AllConv, NiN, VGG-16(C) and the AlexNet network on ImageNet. The curves of fitness value of randomly selected images are drawn with solid gray lines and the average values are highlighted by red dotted lines. To make the comparision clearer, we also put the averages of the three cases (attacking on the original entire image, its foreground and background) together.}.
    \label{fig:fitness}
\end{figure*}

\textbf{Change in fitness values and number of generations.} In order to observe more clearly what happened during the attack, how the fitness values change during the evolution of different networks and different datasets in the case of untargeted attack is illustrated in Fig~\ref{fig:fitness}. More specifically, we randomly selected $15$ images that can be successfully attacked not only on the original entire image but also on its foreground and background for all the three CIFAR-10 networks: AllConv, NiN, VGG-16(C), and selected $10$ ImageNet images for the model AlexNet. Remember that the fitness value is set to be the confidence of the true class of the image, and the goal of the untargeted attack is to minimize this fitness value, as previously mentioned. We also report the average number of generations required for all images successfully attacked on the original image, its foreground and background for all the CIFAR-10 and ImageNet networks in Table~\ref{table:4}.

As shown in Fig~\ref{fig:fitness}, the fitness values can drop off a cliff between certain two generations and decrease steadily at other generations. For the sample that can be successfully attacked both on the original entire image and its foreground and background, the number of generations required to drop the fitness value sufficiently to make the image be misclassified is much less than $100$. The average of fitness decreases monotonically with the number of generations, moreover, the average fitness value drops almost as much for attacking on the foreground as it does for the entire image. However, when attacking on the background of the image, the average fitness value drop is much smaller. The results in Table~\ref{table:4} also show that for all the networks on CIFAR-10 and ImageNet, the average number of generations that required for the successful untargeted attack by modifying one pixel on the background of the image is significantly higher than on the entire image and its foreground. Besides, it can be find that the fitness value decrease less for the AlexNet network compared with the networks on CIFAR-10. This means that under the current settings, the AlexNet network is more difficult to fool.

\begin{table}[!htbp]
\caption{The average number of generations required for the successful untargeted attack when respectively modifying one pixel on the original entire image, its foreground and background for all the three networks on CIFAR-10: AllConv, NiN, VGG-16(C) and the three networks on ImageNet: AlexNet, ResNet-50, VGG-16(I).}
\label{table:4}
\setlength{\tabcolsep}{3.5mm}{
\begin{tabular}{llll}
\toprule          & Original & Foreground & Background \\  \hline
AllConv   & 1.63     & 1.77       & 3.03       \\ \hline
NiN       & 1.36     & 1.48       & 2.45       \\ \hline
VGG-16(C) & 1.45     & 1.69       & 4.70       \\ \midrule[1pt]
AlexNet   & 3.12     & 4.68       & 7.52       \\ \hline
ResNet-50 & 1.00     & 1.00       & 1.20       \\  \hline
VGG-16(I) & 3.00     & 3.50       & 9.50     \\ \bottomrule 
\end{tabular}}
\end{table}

\section{Conclusion}
\label{section:5}
In the present study, we have investigated the differences in adversarial robustness between regions of different semantic importance in an image. We shed light on this issue by perturbing one pixel at a time on the different regions of segmented image to fool the deep neural networks. We used the GrabCut method for CIFAR-10 and U$^{2}$-Net for ImageNet to segment the image into the foreground region which is semantically meaningful and the background region that does not carry much semantic information. We proposed an algorithm improved on the DE algorithm to find the solutions in multiple discontinuous intervals when one-pixel attacks are carried out in part of the image (specifically corresponding to the foreground or the background region of an image in this paper). Moreover, the experimental results on both CIFAR-10 and ImageNet for networks with different architectures demonstrate that there are multiple pixels on an image that can be perturbed to achieve an one-pixel adversarial attack, and attacking on the foreground region is easier to succeed, thus the adversarial robustness is worse than on the background. 

In summary, the work of this paper explores the vulnerabilities in pixel space and reveals the inherent connection between adversarial robustness and semantic information. The algorithm we proposed in this paper implies a kind of new adversarial attack launched in a more extremely limited scenario that has not been considered by previous attacks where the adversary does not know the values of all the pixels in an image when perturbing it. Furthermore, the diversity of the vulnerable points to adversarial attacks on the pixel level suggested in this paper provides insights on the decision boundaries of deep neural networks. Based on the observation of the connection between the adversarial robustness and semantic information of the image, new ideas to improve the adversarial robustness may be inspired. A theoretical study on the above issues is reserved for future work.




%





\ifCLASSOPTIONcaptionsoff
  \newpage
\fi





\bibliographystyle{IEEEtran}
\bibliography{Bibliography}

\vfill


\end{document}